\documentclass[10pt,twocolumn]{article}
\usepackage[utf8]{inputenc}
\usepackage[T1]{fontenc}
\usepackage{amsmath,amssymb,amsfonts}
\usepackage{graphicx}
\usepackage[margin=0.75in]{geometry}
\usepackage{hyperref}
\usepackage{float}
\usepackage{caption}
\usepackage{booktabs}
\usepackage{abstract}
\usepackage{titlesec}
\usepackage{enumitem}
\usepackage{xcolor}
\usepackage{fancyhdr}

\fancypagestyle{firstpage}{
  \fancyhf{}
  \fancyfoot[L]{\footnotesize\textit{© 2026 Ekta Inc. All rights reserved}}
  \fancyfoot[C]{\thepage}

}
\titleformat{\section}{\large\bfseries}{\thesection.}{0.5em}{}
\titleformat{\subsection}{\normalsize\bfseries}{\thesubsection}{0.5em}{}
\titleformat{\subsubsection}{\normalsize\bfseries\itshape}{\thesubsubsection}{0.5em}{}
\DeclareMathOperator*{\argmax}{arg\,max}
\newcommand{\T}{\mathcal{T}}
\newcommand{\F}{\mathcal{F}}
\newcommand{\w}{\mathbf{w}}

\title{\textbf{The Institutional Scaling Law:}\\[4pt]
\large Non-Monotonic Fitness, Capability-Trust Divergence,\\and Symbiogenetic Scaling in Generative AI}

\author{%
  Mark Baciak$^{1}$ and Thomas A.\ Cellucci$^{1}$\\[6pt]
  $^{1}$Ekta Inc.\\
}
\date{}

\begin{document}
\twocolumn[
  \begin{@twocolumnfalse}
    \maketitle
    \thispagestyle{firstpage}
    \begin{abstract}
\noindent Classical scaling laws model AI performance as monotonically improving with model size. We challenge this assumption by deriving the \textit{Institutional Scaling Law}, showing that institutional fitness---jointly measuring capability, trust, affordability, and sovereignty---is \textit{non-monotonic} in model scale, with an environment-dependent optimum $N^*(\varepsilon)$. Our framework extends the Sustainability Index of Han et al.\ (2025) from hardware-level to ecosystem-level analysis, proving that capability and trust formally diverge beyond critical scale (Capability-Trust Divergence). We further derive a Symbiogenetic Scaling correction showing that orchestrated systems of domain-specific models can outperform frontier generalists in their native deployment environments. We contextualize these results within a formal evolutionary taxonomy of generative AI spanning five eras (1943--present), with analysis of frontier lab dynamics, sovereign AI emergence, and post-training alignment evolution.

    \medskip
    \noindent\textbf{Keywords:} generative AI, scaling laws, institutional scaling law, symbiogenetic scaling, capability-trust divergence, multi-agent systems, sovereign AI, punctuated equilibrium
    \end{abstract}
    \vspace{0.5cm}
  \end{@twocolumnfalse}
]

\section{Introduction}

The field of artificial intelligence has undergone a transformation so rapid that conventional narratives of smooth, monotonic progress fail to capture the actual dynamics of change. The emergence of generative AI, powered by the transformer architecture \cite{vaswani2017}, has proceeded not as gradual accumulation but as \textit{punctuated equilibrium} \cite{gould1972}: extended periods of relative stasis interrupted by sudden, transformative leaps.

This paper introduces a formal evolutionary taxonomy for AI development and, more significantly, derives the \textit{Institutional Scaling Law}---a mathematical framework that supersedes classical scaling laws by demonstrating that institutional fitness is non-monotonic in model scale. We formalize our framework by extending the Sustainability Index of Han et al.\ \cite{han2025} from hardware-level model evaluation to an ecosystem-level Institutional Fitness Manifold. From this we derive that orchestrated systems of domain-specific models can outperform frontier generalists in their native environments (Symbiogenetic Scaling, Equation~\ref{eq:agent}).

Our contributions include: (1) the Institutional Fitness Manifold and Institutional Scaling Law formalizing selection pressures driving AI evolution; (2) the Symbiogenetic Scaling correction proving domain-specific model systems can exceed generalist fitness; (3) comprehensive mapping of frontier AI laboratories; (4) analysis of post-training alignment evolution from RLHF through GRPO; (5) introduction of Sovereign AI as a defining selection pressure; and (6) analysis of the DeepSeek Moment---a punctuation event that erased \$589B in market value.

The remainder is organized as follows. Section~2 reviews related work. Section~3 presents our taxonomy, mathematical formalization (\S3.1), and the Institutional Scaling Law (\S3.2). Section~4 analyzes the Generative AI era. Section~5 covers Sovereign AI. Section~6 forecasts future evolution. Section~7 discusses implications and Section~8 concludes.

\section{Related Work}

Loch \& Huberman \cite{loch1999} developed a formal punctuated-equilibrium model of technology diffusion. Valverde \& Sol\'e \cite{valverde2015} applied phylogenetic analysis to programming languages. Kaplan et al.\ \cite{kaplan2020} established power-law scaling relationships for neural language models analogous to allometric scaling in biology. Hoffmann et al.\ \cite{hoffmann2022} refined these with Chinchilla, while Han et al.\ \cite{han2025} demonstrated these linear assumptions break down under quantization, revealing a `quantization trap' with implications for deployment trust.

Han et al.\ formalized evaluation through a three-dimensional Sustainability Index jointly measuring Trust, Economic Efficiency, and Environmental Energy---proving these dimensions can decouple (Amortization-Trust Decoupling). We extend their framework in \S3.1--3.2 by adding sovereign compliance, introducing environment-dependence, and deriving a new scaling law. Lu et al.\ \cite{lu2026} demonstrated dynamic topology routing among specialized LLM agents outperforms fixed patterns (+6.2\% avg.), supporting our Symbiogenetic Scaling correction. Wang et al.\ \cite{wang2024alignment} survey alignment evolution; Chen et al.\ \cite{chen2025agents} document self-evolving agents. None adopt the integrated evolutionary framework or formal scaling law we propose.

\section{A Formal Evolutionary Taxonomy of AI}

We propose a hierarchical taxonomy modeled on the geological timescale (Eon $>$ Era $>$ Epoch). Each of five eras is defined by a dominant computational paradigm, with boundaries marked by phase transition events. Within the current era (Generative AI), we identify four epochs bounded by punctuation events: GPT-3 (June 2020), ChatGPT (November 2022), and OpenAI o1 (September 2024). Table~\ref{tab:taxonomy} summarizes the full taxonomy.

\begin{table}[t]
\centering
\caption{Evolutionary Taxonomy of AI Development}
\label{tab:taxonomy}
\small
\begin{tabular}{@{}p{2.2cm}p{1.1cm}p{3.8cm}@{}}
\toprule
\textbf{Era} & \textbf{Period} & \textbf{Phase Transition} \\
\midrule
1. Abiogenesis & 1943--56 & Dartmouth Conference \\
2. Paleozoic & 1956--86 & Expert system collapse \\
3. Mesozoic & 1986--12 & AlexNet/ImageNet \\
4. Cenozoic & 2012--17 & Transformer (2017) \\
5. Generative AI & 2017-- & \textit{(current era)} \\
\bottomrule
\end{tabular}
\end{table}

\subsection{Mathematical Formalization: The Institutional Fitness Manifold}
\label{sec:manifold}

We build on Han et al.'s \cite{han2025} Sustainability Index framework, extending it from hardware-level to ecosystem-level analysis.

\textbf{Definition 1 (Institutional Fitness Vector).} For any AI configuration $\theta \in \Theta$ deployed in environment $\varepsilon \in \mathcal{E}$:
\begin{equation}
\label{eq:fitness_vec}
f(\theta, \varepsilon) = \bigl(C(\theta),\; \T(\theta,\varepsilon),\; A(\theta),\; \Sigma(\theta,\varepsilon)\bigr)^\top \in [0,1]^4
\end{equation}
where $C$ is capability, $\T$ is institutional trust (environment-dependent), $A$ is affordability, and $\Sigma$ is sovereignty compliance. This extends Han et al.'s three-dimensional vector to four dimensions with environment-dependence.

\textbf{Definition 2 (Scalar Fitness).} The scalar fitness in environment $\varepsilon$ is:
\begin{equation}
\label{eq:scalar}
F(\theta, \varepsilon) = \w(\varepsilon)^\top \cdot f(\theta, \varepsilon), \quad \sum_i w_i(\varepsilon) = 1
\end{equation}
where the weight vector $\w(\varepsilon)$ varies by deployment context.

\textbf{Theorem 1 (Capability-Trust Divergence).}
\begin{equation}
\label{eq:divergence}
\frac{\partial F}{\partial N} = w_C \frac{\partial C}{\partial N} + w_{\T} \frac{\partial \T}{\partial N} + w_A \frac{\partial A}{\partial N} + w_\Sigma \frac{\partial \Sigma}{\partial N}
\end{equation}
Since $\partial C/\partial N > 0$ (Kaplan scaling with $\alpha \approx 0.076$) but $\partial \T/\partial N < 0$ beyond critical threshold $N^*$ (larger models are harder to audit), the gradient $\partial F/\partial N$ flips sign---producing a \textit{Capability-Trust Divergence} directly analogous to Han et al.'s Scaling Law Divergence ($\partial \text{SI}/\partial p > 0$), but at the ecosystem level.

\textbf{Theorem 2 (Sequential Trust Degradation).}
\begin{equation}
\label{eq:trust_deg}
\T_{\text{inst}}(K) = \prod_{k=1}^{K}(1-\varepsilon_k) \approx e^{-\sum \varepsilon_k}, \quad \frac{\partial \T_{\text{inst}}}{\partial K} < 0
\end{equation}
structurally identical to Han et al.'s multi-hop fragility formalization. Each deployment context is a `hop' across which trust-eroding incidents compound.

\textbf{Proposition 1 (Speciation via Environmental Isolation).} Let $\theta^*(\varepsilon) = \argmax_\theta F(\theta, \varepsilon)$. If two environments $\varepsilon_1, \varepsilon_2$ have sufficiently different weight vectors:
\begin{equation}
\label{eq:speciation}
\| \theta^*(\varepsilon_1) - \theta^*(\varepsilon_2) \| \geq \kappa \cdot \| \w(\varepsilon_1) - \w(\varepsilon_2) \|, \quad \kappa > 0
\end{equation}
formalizing that sovereign AI is a \textit{mathematical necessity} from divergent fitness landscapes.

\textbf{Definition 3 (Phase Transition Detection).}
\begin{equation}
\label{eq:phase}
\left.\frac{d}{dt} H(\Psi(t))\right|_{t=t^*} > \lambda_{\text{crit}}, \quad H(\Psi) = -\sum_i p_i \ln p_i
\end{equation}
where $H(\Psi)$ is the Shannon entropy of the configuration-deployment distribution. Spikes in $dH/dt$ indicate punctuated equilibrium events.

\textit{Remark.} The framework admits empirical testing: weight vectors $\w(\varepsilon)$ can be estimated from procurement data; trust degradation from safety incidents; entropy rate from deployment surveys.

\subsection{The Institutional Scaling Law}
\label{sec:scaling_law}

Classical scaling laws model loss as a monotonically improving function of scale. Han et al.\ \cite{han2025} showed this breaks at the hardware level. We show an analogous breakdown at the \textit{ecosystem level}: institutional fitness is \textit{non-monotonic} in model scale.

\textbf{Proposition 2 (The Institutional Scaling Law).}
\begin{align}
\label{eq:inst_scaling}
\F(N&,p,K,\varepsilon) = w_C\!\left[1 - \!\left(\frac{N_c}{N}\right)^{\!\alpha}\right]\nonumber\\
&+ w_{\T}\!\left[\T_0 e^{-\beta N^\gamma}\right] + w_A\!\left[\!\left(\frac{N_r}{N}\right)^{\!\delta}\!\Phi(p)\right] + w_\Sigma \sigma(\varepsilon)
\end{align}
where: \textit{Capability} $C(\theta) = 1 - (N_c/N)^\alpha$ follows Kaplan's power law ($\alpha \approx 0.076$); \textit{Trust} $\T(\theta, \varepsilon) = \T_0 \cdot e^{-\beta N^\gamma}$ decays exponentially beyond critical scale; \textit{Affordability} $A(\theta) = (N_r/N)^\delta \cdot \Phi(p)$ captures cost-per-query modulated by quantization efficiency; and \textit{Sovereignty} $\Sigma(\theta, \varepsilon) = \sigma(\varepsilon)$ is the environment-specific compliance index.

The \textit{Phase Boundary} is the first-order optimality condition:
\begin{align}
\label{eq:phase_boundary}
N^*(\varepsilon)&:\; \left.\frac{\partial \F}{\partial N}\right|_{N^*} \!= 0 \;\Rightarrow\nonumber\\
w_C \alpha \frac{N_c^\alpha}{N^{*(\alpha+1)}} &= w_{\T} \beta\gamma \T_0 N^{*(\gamma-1)} e^{-\beta N^{*\gamma}}\nonumber\\
&\quad + w_A \delta \frac{N_r^\delta}{N^{*(\delta+1)}}\Phi(p)
\end{align}
Below $N^*$, capability gains dominate. Above $N^*$, trust and cost penalties dominate and bigger is \textit{worse}. $N^*(\varepsilon)$ varies by environment: a Silicon Valley startup may find $N^* \approx 140$B, while an EU regulated institution finds $N^* \approx 45$B.

The quantization-trust interaction directly incorporates Han et al.'s framework:
\begin{equation}
\label{eq:quant_trust}
\Phi(p) = \min\!\left(1,\; \frac{\log(1 + \chi_{\text{ref}})}{\log(1 + \chi(p))}\right), \quad \chi(p) = E_q(p) \cdot \gamma_{\text{grid}}
\end{equation}

\begin{figure*}[t]
\centering
\includegraphics[width=0.85\textwidth]{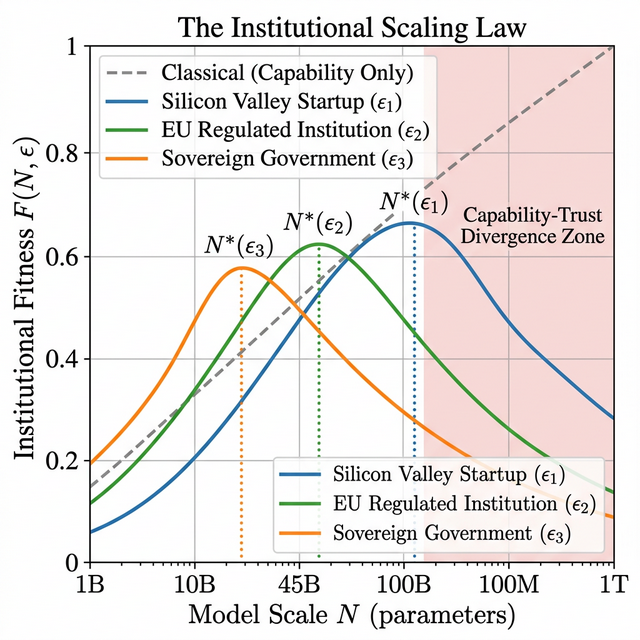}
\caption{The Institutional Scaling Law. Institutional fitness $\F(N, \varepsilon)$ is non-monotonic: each deployment environment has a distinct optimal scale $N^*(\varepsilon)$. The dashed line shows the classical capability-only view. The shaded region marks the Capability-Trust Divergence Zone.}
\label{fig:scaling}
\end{figure*}

\subsubsection{Symbiogenetic Scaling: The Multi-Agent System Correction}

The Symbiogenesis epoch (Section~4.4) is characterized by the fusion of specialized systems. This introduces a qualitatively different scaling regime, formalized through a multi-agent topology correction motivated by Lu et al.\ \cite{lu2026}:
\begin{equation}
\label{eq:agent}
\F_{\text{agent}}(N, K, G) = \F(N, p, K, \varepsilon) \cdot \left[1 + \eta \cdot \frac{\rho(G)}{\sqrt{K}}\right]
\end{equation}
where $K$ is the number of agents, $G$ the communication graph, $\rho(G) = |E_{\text{eff}}|/K(K\!-\!1)$ is the effective communication density, and $\eta > 0$ is orchestration efficiency. The key insight: \textit{domain-specific models tightly coupled to specific tools, trained on system schema and data, and coordinated through adaptive topology routing, can collectively exceed the fitness of a generalist frontier model that has never encountered the deployment environment's tools or constraints.}

The \textit{Convergence-Orchestration Threshold} formalizes when orchestration outweighs scaling:
\begin{equation}
\label{eq:convergence}
N_{\text{conv}}:\; \left.\frac{\partial C}{\partial N}\right|_{N_{\text{conv}}} < \mu \;\Longrightarrow\; \frac{\partial \F_{\text{agent}}}{\partial G} > \frac{\partial \F_{\text{agent}}}{\partial N}
\end{equation}
Once capability saturates ($\partial C/\partial N < \mu$), investment in orchestration topology dominates investment in scale. This is precisely \textit{symbiogenesis} \cite{margulis1967}: the mitochondrion did not outcompete the cell---it merged with it.

\textbf{Corollary (Scaling Law Inversion).} For environments where $w_{\T}(\varepsilon) + w_\Sigma(\varepsilon) > 0.5$, there exists a system of $K$ domain-specific models with $N_i \ll N_{\text{frontier}}$ such that:
$$\F_{\text{agent}}\!\left(\textstyle\sum N_i, K, G, \varepsilon\right) > \F(N_{\text{frontier}}, p, 1, \varepsilon)$$
The classical scaling law says bigger is always better. The Institutional Scaling Law says \textit{better-adapted is always better}.

\section{The Generative AI Era: Deep Analysis}

\subsection{Epoch I: Morphogenesis (2017--2020)}

The transformer \cite{vaswani2017} initiated the Generative AI era. GPT-1 \cite{radford2018} (117M), BERT \cite{devlin2018} (340M), GPT-2 \cite{radford2019} (1.5B), and GPT-3 \cite{brown2020} (175B) demonstrated emergent few-shot learning. Kaplan et al.\ \cite{kaplan2020} established neural scaling laws:
\begin{equation}
\label{eq:kaplan}
L(N) = \left(\frac{N_c}{N}\right)^{\!\alpha_N}\!, \quad \alpha_N \approx 0.076
\end{equation}
converting AI development from empirical art into predictive science.

\subsection{Epoch II: Adaptive Radiation (2020--2022)}

An explosive diversification followed: DALL$\cdot$E \cite{ramesh2021}, Codex \cite{chen2021code}, PaLM \cite{chowdhery2022}, Stable Diffusion \cite{rombach2022}, AlphaFold \cite{jumper2021}. Hoffmann et al.\ \cite{hoffmann2022} refined scaling with Chinchilla, demonstrating many models were undertrained. By 2022, transformer-based systems had colonized every major computational modality.

\subsection{Epoch III: The Great Expansion (Nov 2022--2024)}

ChatGPT reached 100M users in two months---the fastest consumer adoption in history. RLHF \cite{ouyang2022} demonstrated that a 1.3B InstructGPT could be preferred over 175B GPT-3 on instruction-following, challenging monotonic capability-scale assumptions. GPT-4 \cite{openai2023} demonstrated multimodal reasoning. Enterprise adoption surged from 6\% (2023) to 30\% (2025).

\subsection{Epoch IV: Symbiogenesis (Sep 2024--Present)}

OpenAI's o1 (September 2024) demonstrated extended chain-of-thought reasoning. The epoch is characterized by \textit{symbiogenesis} \cite{margulis1967}: language models, code interpreters, search, memory, and tools fusing into integrated agents. By July 2025, ChatGPT reached 700M weekly active users. The DeepSeek Moment (\S4.5.1) erased \$589B in NVIDIA market cap and challenged hardware-dependent scaling assumptions.

\subsection{Frontier Lab Ecosystem}

The competitive ecosystem spans the US (OpenAI, Anthropic, Meta, xAI), China (DeepSeek, Alibaba/Qwen, Baidu, Moonshot, MiniMax, 01.AI), Europe (Mistral, Black Forest Labs), Canada (Cohere), and Israel (AI21 Labs). Table~\ref{tab:labs} summarizes key laboratories.

\begin{table*}[t]
\centering
\caption{Major Frontier AI Laboratories (as of February 2026)}
\label{tab:labs}
\small
\begin{tabular}{@{}llllp{6.5cm}@{}}
\toprule
\textbf{Laboratory} & \textbf{Country} & \textbf{Entry} & \textbf{First Model} & \textbf{Key Contributions} \\
\midrule
OpenAI & USA & Jun 2018 & GPT-1 (117M) & GPT series through GPT-5, o-series reasoning, DALL$\cdot$E, Stargate (\$500B) \\
Google DeepMind & UK/USA & Oct 2018 & BERT (340M) & Transformer co-origin, PaLM, Gemini series, AlphaFold \\
Anthropic & USA & Mar 2023 & Claude 1.0 & Constitutional AI (RLAIF), Claude series through Opus 4.6 \\
Meta AI (FAIR) & USA & Feb 2023 & LLaMA (65B) & Llama open-weight series, PyTorch, open-source leadership \\
Mistral AI & France & Sep 2023 & Mistral 7B & Mixtral MoE, European sovereignty, Apache 2.0 \\
DeepSeek & China & Nov 2023 & DeepSeek Coder & R1 reasoning (Jan 2025), V3 MoE, cost-efficiency breakthroughs \\
Alibaba (Qwen) & China & Aug 2023 & Qwen-7B & Qwen series through Qwen 3, multilingual, open-weight \\
xAI & USA & Nov 2023 & Grok-1 & Grok series, Colossus 100K-GPU cluster \\
Moonshot AI & China & Oct 2023 & Kimi Chat & Kimi K2 Thinking, outperforming GPT-5 on select benchmarks \\
MiniMax & China & Dec 2023 & abab5.5 & M2 record open-model scores (2025) \\
\bottomrule
\end{tabular}
\end{table*}

\subsubsection{The DeepSeek Moment: A Punctuation Event}

On January 20, 2025, DeepSeek released R1, matching OpenAI's o1 at reportedly under \$6M training cost using export-compliant H800 GPUs. NVIDIA lost \$589B in a single day---the largest single-day value loss in stock market history. This punctuation event demonstrated that algorithmic efficiency could substitute for raw compute, validated open-source release, and forced industry-wide strategic recalibration.

\subsubsection{The Lunar New Year Effect}

By 2026, Chinese firms time releases to the Spring Festival. The six weeks preceding this paper saw: Alibaba's Qwen 3.5, ByteDance's Doubao 2.0, Zhipu AI's GLM-5 (trained entirely on Huawei Ascend), and multiple Western releases (Claude Opus 4.6, GPT-5.3 Codex, Gemini 3.1 Pro). The ecosystem has developed periodic bursts consistent with punctuated equilibrium.

\subsection{Post-Training Alignment Evolution}

Post-training methods have undergone six paradigm shifts in four years:

\textbf{RLHF + PPO} \cite{ouyang2022}: Three-model pipeline (policy, reward, value). Enabled ChatGPT.
\begin{equation}
\label{eq:rlhf}
\pi^* = \argmax_\pi \mathbb{E}_{x,y\sim\pi}\bigl[r_\varphi(x,y)\bigr] - \beta \cdot D_{\text{KL}}(\pi \| \pi_{\text{ref}})
\end{equation}

\textbf{DPO} \cite{rafailov2023}: Eliminated reward model via reparameterization.
{\small
\begin{equation}
\label{eq:dpo}
\mathcal{L}_{\text{DPO}} = -\mathbb{E}\!\left[\log \sigma\!\left(\beta \log \frac{\pi_\theta(y_w|x)}{\pi_{\text{ref}}(y_w|x)} - \beta \log \frac{\pi_\theta(y_l|x)}{\pi_{\text{ref}}(y_l|x)}\right)\right]
\end{equation}
}

\textbf{GRPO} \cite{shao2024,deepseek2025}: Eliminated both reward model and critic. For each prompt $x$, sample $G$ responses, score by verifiable reward, normalize within group:
{\small
\begin{equation}
\label{eq:grpo}
\mathcal{L}_{\text{GRPO}} = -\frac{1}{G}\sum_{i=1}^{G}\min\!\left(\frac{\pi_\theta(o_i|x)}{\pi_{\text{old}}(o_i|x)}\hat{A}_i,\, \text{clip}\!\left(\frac{\pi_\theta}{\pi_{\text{old}}}, 1\!\pm\!\epsilon\right)\!\hat{A}_i\right)
\end{equation}
}
GRPO's reliance on verifiable rewards proved decisive in DeepSeek-R1's training, enabling sophisticated reasoning through pure RL.

Han et al.\ \cite{han2025} demonstrated the quantization trap---compression paradoxically increases energy while degrading reasoning:
\begin{equation}
\label{eq:qtrap}
E_q(b, d) \propto b^{-1} \cdot d \cdot \gamma_{\text{grid}}, \quad \gamma_{\text{grid}} \gg 1 \;\text{for low}\; b
\end{equation}
This creates a structural tension: the capability that makes GRPO-trained models powerful (multi-step reasoning) is precisely what quantization degrades most.

\section{The Rise of Sovereign AI}

Sovereign AI demands control of the entire cognitive stack---from training data to model weights to policy alignment. We identify four dimensions: (1) \textit{Training Data Sovereignty}; (2) \textit{Model Sovereignty}; (3) \textit{Infrastructure Sovereignty}; (4) \textit{Interaction Sovereignty}---ensuring prompts, queries, and outputs remain within sovereign boundaries. McKinsey projects a \$600B sovereign AI market by 2030.

\textbf{National Programs.} The US launched the \$500B Stargate Project. China matched combined US/UK/EU AI research output in 2024 with \$140B+ state investment. The EU launched OpenEuroLLM across 24 languages. The UAE developed Falcon and Jais via G42/MBZUAI. India's IndiaAI Mission (INR 10,300+ crore) hosted the first Global South AI summit (February 2026). Sarvam AI launched 30B/105B MoE models for India's multilingual landscape.

\textbf{Davos 2026.} The WEF Annual Meeting (January 2026) featured sovereign AI as the dominant technology theme. WEF/Bain projected \$1.5T for applications and \$400B for infrastructure annually by 2030.

\textbf{India AI Impact Summit 2026.} The first Global South AI summit drew 100+ countries, 20 heads of state. Investment commitments exceeded \$200B. The White House rejected global AI governance outright, while middle powers increasingly seek to build independent capabilities---precisely the selection pressure driving speciation.

From an evolutionary perspective, sovereign AI introduces a new \textit{ecological niche} driving adaptive radiation. Models must differentiate on cultural attunement, linguistic coverage, regulatory compliance, and data provenance---mirroring biological \textit{character displacement}.

\section{Forecasting Future Evolution}

\subsection{The Next Epoch: Noogenesis ($\sim$2026--2030?)}

Key indicators include: autonomous scientific discovery (a 27B Gemma-based model generated a validated cancer hypothesis in October 2025); self-improving architectures (GPT-5 Codex works independently for 7+ hours); sovereign AI ecosystem fragmentation; cost collapse (30$\times$ reduction in three years); and domain-specific model compression (2B--10B models exceeding generalist performance on domain tasks while sidestepping the quantization trap by running at full precision on commodity hardware).

\subsection{The Next Era: The Post-Transformer (?)}

A new era requires a \textit{discontinuous substrate change}. Candidates include: state-space models (Mamba), neuromorphic computing, and quantum-enhanced architectures. Each represents evolutionary innovation \textit{within} the current lineage, not a new class of organism.

\subsection{The Latent Capability Paradox}

Frontier models exhibit persuasive strategies, implicit persona adaptation, and linguistic influence patterns that resist comprehensive characterization. The convergence of expanding latent capabilities and eroding safety infrastructure is producing a trust deficit that itself becomes a \textit{selection pressure}: nations seek auditable models (Sovereign AI), and the market shifts toward smaller, verifiable, domain-specific alternatives. The Institutional Fitness Manifold provides the formal structure: Capability-Trust Divergence (Theorem~1, Eq.~\ref{eq:divergence}) predicts the sign-flip; Sequential Trust Degradation (Theorem~2, Eq.~\ref{eq:trust_deg}) quantifies compounding erosion; and Speciation (Eq.~\ref{eq:speciation}) demonstrates divergent environments necessitate divergent optima.

\section{Discussion}

The evidence strongly supports punctuated equilibrium in AI evolution. Era-level transitions---Dartmouth (1956), backpropagation (1986), AlexNet (2012), transformer (2017)---each produced discontinuous, irreversible shifts. Epoch-level events---GPT-3 (2020), ChatGPT (2022), o1 (2024)---reorganized the competitive landscape at accelerating frequency.

The MIT NANDA \textit{State of AI in Business 2025} report found 95\% of enterprise AI pilots produce zero measurable P\&L impact. The ``GenAI Divide'' reveals that institutional absorption lags far behind technical innovation, with a ``shadow AI economy'' where 90\% of employees use personal AI tools while official initiatives stall.

The Institutional Scaling Law (Eq.~\ref{eq:inst_scaling}) provides the formal structure for the emerging bifurcation between capability and trust. At $N^*(\varepsilon)$, capability gain exactly offsets trust/cost penalties. The Symbiogenetic Scaling correction (Eq.~\ref{eq:agent}) and Convergence-Orchestration Threshold (Eq.~\ref{eq:convergence}) formalize the alternative: orchestrated domain-specific models coupled to institutional tools and data can exceed any individual frontier model's institutional fitness.

\textbf{Limitations.} Evolutionary metaphors are imperfect. Technological evolution involves deliberate design and economic selection pressures differing from natural selection. Our taxonomy is retrospective; era boundaries involve judgment.

\section{Conclusion}

This paper has introduced the Institutional Scaling Law, demonstrating that institutional fitness is a non-monotonic function of model scale with environment-dependent optimum $N^*(\varepsilon)$. The Capability-Trust Divergence theorem provides a formal mechanism for why scaling up can reduce institutional fitness; Sequential Trust Degradation explains compounding trust erosion; the Speciation Proposition derives ecosystem fragmentation from environmental heterogeneity; and the Symbiogenetic Scaling correction formalizes the prediction that the next phase transition will be driven not by larger models but by better-orchestrated systems of domain-specific models that fuse into composite intelligences adapted to specific institutional niches.

The classical scaling law says bigger is always better. The Institutional Scaling Law says \textit{better-adapted is always better}---and at the ecosystem level, adaptation increasingly means orchestrated specialization rather than undifferentiated scale.

\section*{Acknowledgements}

The authors gratefully acknowledge Dr.\ Gunnar E.\ Carlsson (Stanford University), Dr.\ Anupam Chattopadhyay (Nanyang Technological University, Singapore), and Cardinal Peter Turkson (Pontifical Academy of Sciences) for their valuable contributions and support.

\bibliographystyle{plain}

\end{document}